\documentclass[10pt,journal,compsoc]{IEEEtran}
\PassOptionsToPackage{dvipsnames}{xcolor}

\ifCLASSOPTIONcompsoc
  \usepackage[nocompress]{cite}
\else
  \usepackage{cite}
\fi

\ifCLASSINFOpdf
  \usepackage[pdftex]{graphicx}
  \DeclareGraphicsExtensions{.pdf,.jpeg,.png}
\else
  \usepackage[dvips]{graphicx}
  \DeclareGraphicsExtensions{.eps}
\fi

\ifCLASSOPTIONcompsoc
  \usepackage[caption=false,font=footnotesize,labelfont=sf,textfont=sf]{subfig}
\else
  \usepackage[caption=false,font=footnotesize]{subfig}
\fi

\usepackage{ragged2e}
\usepackage{lipsum}
\usepackage{comment}
\usepackage{color,soul}
\usepackage{booktabs}
\usepackage{multirow}
\usepackage{tabularx}
\usepackage{amsmath}
\usepackage{bm}
\PassOptionsToPackage{hyphens}{url}%\usepackage{hyperref}
\usepackage{hyperref}
\usepackage{mathtools}

\usepackage{tikz,tikz-3dplot}
\usetikzlibrary{calc}
\usetikzlibrary{shapes,positioning,intersections,quotes}
\usepackage{graphicx}
\usepackage{footmisc}
\usepackage{booktabs}
\usepackage{amsmath}
\usepackage{amsfonts}
\usepackage{amsbsy}
\usepackage{placeins}
\usepackage{siunitx}
\usepackage{bm}
\usepackage[utf8]{inputenc} % allow utf-8 input
\usepackage[T1]{fontenc}    % use 8-bit T1 fonts
\usepackage{url}            % simple URL typesetting
\usepackage{nicefrac}       % compact symbols for 1/2, etc.
\usepackage{multirow}
\usepackage{mathtools}
\usepackage{xparse}
\usepackage{empheq}

\usepackage[switch]{lineno}
\usepackage{centernot}
\usepackage{graphicx}
\usepackage{footmisc}
\usepackage{booktabs}
\usepackage{amsmath}
\usepackage{amsfonts}
\usepackage{amsbsy}
\usepackage{placeins}
\usepackage{siunitx}
\usepackage{bm}
\usepackage{multirow}
\usepackage{mathtools}
\usepackage{xparse}
\DeclarePairedDelimiter\abs{\lvert}{\rvert}
\DeclarePairedDelimiter\norm{\lVert}{\rVert}

\usepackage{empheq}
\usepackage{tabularx} 
\newcolumntype{C}{>{\centering\arraybackslash}X}
\usepackage{subfig}
\usepackage{centernot}
\usepackage{ragged2e}

\newcommand\blfootnote[1]{%
  \begingroup
  \renewcommand\thefootnote{}\footnote{#1}%
  \addtocounter{footnote}{-1}%
  \endgroup
}

\usepackage{algorithm}
\usepackage{listings}
\usepackage{cuted}
\begin{document}

%\title{Enhanced Isotropy Maximization Loss:\\Seamless and High-Performance Out-of-Distribution Detection\\Simply Replacing the SoftMax Loss}
\title{Enhanced Isotropy Maximization Loss:\\Seamless and High-Performance\\Out-of-Distribution Detection\\Simply Replacing the SoftMax Loss}

\author{
David~Mac\^edo,~\IEEEmembership{Member,~IEEE,}
and~Teresa~Ludermir,~\IEEEmembership{Senior Member,~IEEE}% <-this % stops a space
\IEEEcompsocitemizethanks{
\IEEEcompsocthanksitem David~Mac\^edo and Teresa~Ludermir are with the Centro de Inform\'atica, Universidade Federal de Pernambuco, Brazil. E-mail: dlm@cin.ufpe.br. David~Mac\^edo was with Montreal Institute for Learning Algorithms (MILA), Université de Montréal (UdeM), Quebec, Canada.% David~Mac\^edo was with Montreal Institute for Learning Algorithms, University of Montreal, Quebec, Canada.
%David~Mac\^edo was with Montreal Institute for Learning Algorithms, University of Montreal, Quebec, Canada.
\protect\\
% note need leading \protect in front of \\ to obtain a newline within \thanks as
%E-mail: see dlm@cin.ufpe.br
%\IEEEcompsocthanksitem David~Mac\^edo was with Montreal Institute for Learning Algorithms, University of Montreal, Quebec, Canada.
}% <-this % stops an unwanted space
%\thanks{Manuscript received August 1, 2020; revised November 1, 2020.}
}

\begin{comment}
\markboth{IEEE TRANSACTIONS ON PATTERN ANALYSIS AND MACHINE INTELLIGENCE}{Mac\^edo~\MakeLowercase{\textit{et al.}}: Enhanced Isotropy Maximization Loss}
\end{comment}

%\maketitle % COMMENT FOR TPAMI

\IEEEtitleabstractindextext{% UNCOMMENT FOR TPAMI
\begin{abstract}
Current out-of-distribution detection approaches usually present special requirements (e.g., collecting outlier data and hyperparameter validation) and produce side effects (e.g., classification accuracy drop and slow/inefficient inferences). Recently, entropic out-of-distribution detection has been proposed as a seamless approach (i.e., a solution that avoids all previously mentioned drawbacks). The entropic out-of-distribution detection solution uses the IsoMax loss for training and the entropic score for out-of-distribution detection. The IsoMax loss works as a drop-in replacement of the SoftMax loss (i.e., the combination of the output linear layer, the SoftMax activation, and the cross-entropy loss) because swapping the SoftMax loss with the IsoMax loss requires no changes in the model's architecture or training procedures/hyperparameters. In this paper, we perform what we call an isometrization of the distances used in the IsoMax loss. Additionally, we propose replacing the entropic score with the minimum distance score. Experiments showed that these modifications significantly increase out-of-distribution detection performance while keeping the solution seamless. Besides being competitive with or outperforming all major current approaches, the proposed solution avoids all their current limitations, in addition to being much easier to use because only a simple loss replacement for training the neural network is required. The code to replace the SoftMax loss with the IsoMax+ loss and reproduce the results is available at \url{https://github.com/dlmacedo/entropic-out-of-distribution-detection}.
\end{abstract}

}% UNCOMMENT FOR TPAMI

\maketitle % UNCOMMENT FOR TPAMI

\IEEEdisplaynontitleabstractindextext % UNCOMMENT FOR TPAMI
\IEEEpeerreviewmaketitle
\IEEEraisesectionheading{
\section{Introduction}\label{sec:introduction}
}% UNCOMMENT FOR TPAMI

\IEEEPARstart{N}{eural} networks have been used in classification tasks in many real-world applications \cite{devries2018leveraging}. In such cases, the system must typically identify whether a given input belongs to any of the classes on which it was trained. \cite{hendrycks2017baseline} called this capability out-of-distribution (OOD) detection and proposed datasets and metrics to allow standardized performance evaluation and comparison. However, current OOD detection solutions still have limitations (e.g., special requirements and side effects) that prevent more general use of OOD detection capabilities in practical real-world applications \cite{macdo2019isotropic} (Table~\ref{tab:overview}).

\begin{table*}%[!ht]
%\small
%\footnotesize
%\scriptsize
%\setlength{\tabcolsep}{2pt}
\caption{\textbf{Out-of-distribution detection approaches: special requirements and side effects.}}
\vskip -0.25cm
%\vskip -0.2cm
%\vspace{-0.1in}
\label{tab:overview}
\centering
%\begin{tabularx}{\textwidth}{clCCC}
\begin{tabularx}{\textwidth}{l|CC|CC}
\toprule
\multirow{3}{*}{Approach}
& \multicolumn{2}{c|}{Special Requirement}
& \multicolumn{2}{c}{Side Effect}\\
\cmidrule{2-5}
%& Hyperparameter & Outlier & Slow/Inefficient & Classification\\
%& Tuning & Data & Inference & Accuracy Drop\\
& Hyperparameter & Additional & Inefficient Inference & Classification\\
& Tuning & Data & or OOD Detection & Accuracy Drop\\
\midrule
%Inference Input Preprocessing: & & & &\\
ODIN \cite{liang2018enhancing} & \color{red}Required & \color{blue}Not Required & \color{red}Present & \color{blue}Not Present\\
\midrule%{2-5}
Mahalanobis \cite{lee2018simple} & \color{red}Required & \color{blue}Not Required & \color{red}Present & \color{blue}Not Present\\
\midrule%{2-5}
ACET \cite{Hein2018WhyRN} & \color{red}Required & \color{blue}Not Required & \color{blue}Not Present & \color{red}Present\\
\midrule%{2-5}
Outlier Exposure \cite{hendrycks2018deep} & \color{blue}Not Required & \color{red}Required & \color{blue}Not Present & \color{blue}Not Present\\
\midrule%{2-5}
Generalized ODIN \cite{Hsu2020GeneralizedOD} & \color{red}Required & \color{blue}Not Required & \color{red}Present & \color{red}Present\\
\midrule%{2-5}
Gram Matrices \cite{Sastry2019DetectingOE} & \color{blue}Not Required & \color{blue}Not Required & \color{red}Present & \color{blue}Not Present\\
\midrule%{2-5}
Scaled Cosine \cite{techapanurak2019hyperparameterfree} & \color{blue}Not Required & \color{blue}Not Required & \color{blue}Not Present & \color{red}Present\\
\midrule%{2-5}
Energy-based \cite{DBLP:journals/corr/abs-2010-03759} & \color{red}Required & \color{red}Required & \color{blue}Not Present & \color{blue}Not Present\\
\midrule%{2-5}
\bf{Entropic (Seamless) \cite{macdo2019isotropic,DBLP:journals/corr/abs-2006.04005}} & \bf{\color{blue}Not} & \bf{\color{blue}Not} & \bf{\color{blue}Not} & \bf{\color{blue}Not}\\
\bf{IsoMax + Entropic Score} & \bf{\color{blue}Required} & \bf{\color{blue}Required} & \bf{\color{blue}Present} & \bf{\color{blue}Present}\\
\midrule%{2-5}
\bf{Entropic (Seamless) [ours]} & \bf{\color{blue}Not} & \bf{\color{blue}Not} & \bf{\color{blue}Not} & \bf{\color{blue}Not}\\
\bf{Enhanced IsoMax + Distance Score} & \bf{\color{blue}Required} & \bf{\color{blue}Required} & \bf{\color{blue}Present} & \bf{\color{blue}Present}\\
\bottomrule
\end{tabularx}
%\vskip 0.1cm
\end{table*}

First, OOD detection solutions commonly present hyperparameters that usually presume access to out-of-distribution samples to be defined \cite{liang2018enhancing, Liang2017PrincipledDO, lee2018simple, lee2018training, DeVries2018LearningNetworks}. A consequence of presuming access to OOD samples to validate hyperparameters and using the same distribution to evaluate OOD detection results is producing overestimated performance estimations \cite{shafaei2018biased}. To avoid unrealistic access to OOD samples and overestimated performance, \cite{lee2018simple} proposed validating hyperparameters using adversarial samples. However, this requires the generation of adversarial examples. This procedure also requires determining hyperparameters (e.g., maximum adversarial perturbation) that are typically unknown when dealing with novel datasets. Similar arguments hold for solutions based on adversarial training \cite{Hein2018WhyRN, lakshminarayanan2017simple, li2018anomaly, kliger2018novelty, lee2018training}, which also result in higher training time. Approaches based on the generation of adversarial examples or adversarial training may also have limited scalability when dealing with large images (e.g., ImageNet \cite{Deng2009ImageNetDatabase}).

Many solutions make use of the so-called \emph{input preprocessing} technique introduced in ODIN \cite{liang2018enhancing}. However, the use of the mentioned technique \emph{increases at least four times the inference delay and power consumption} \cite{macdo2019isotropic, DBLP:journals/corr/abs-2006.04005} because a combination of a first forward pass, backpropagation operation, and second forward pass is required \cite{liang2018enhancing, lee2018simple, Hsu2020GeneralizedOD, DeVries2018LearningNetworks} for a single useful inference. Actually, approaches that may be applied directly to pretrained models and completely avoid training or fine-tuning the model \cite{liang2018enhancing, lee2018simple, Sastry2019DetectingOE} typically produce inefficient inferences and/or additional computational complexity to perform OOD detection \cite[Section IV, D]{DBLP:journals/corr/abs-2006.04005}. From a practical perspective, this is a drawback because inferences may be performed thousands or millions of times in the field. Thus, such approaches may be prohibitive (not sustainable) from environmental \cite{Schwartz2019GreenA} %\footnote{\url{https://www.youtube.com/watch?v=KnOpWgUCtaM}}
and real-world cost-based perspectives.

Another harmful and common side effect is the so-called \emph{classification accuracy drop} \cite{techapanurak2019hyperparameterfree, Hsu2020GeneralizedOD}. In such cases, higher OOD detection performance is achieved at the expense of a drop in the classification accuracy compared with models trained using the usual SoftMax loss (i.e., the combination of the output linear layer, the SoftMax activation, and the cross-entropy loss \cite{liu2016large})\footnote{In this paper, we consider that an approach does not present classification accuracy drop if it never presents a classification accuracy more than one percent lower than the correspondent softmax-loss-trained model.}. For example, in some situations, the solution proposed in \cite{techapanurak2019hyperparameterfree} produces a classification accuracy drop of more than two percent when compared with SoftMax loss training of the same neural network. From a practical perspective, this situation is not desirable because the detection of out-of-distribution samples may be a rare event, and classification is usually the primary aim of the designed system \cite{carlini2019evaluating}. The proposal also requires changing the training of the last layer by removing its weight decay to work properly. Therefore, this solution does not work as a SoftMax loss drop-in replacement.

\begin{algorithm*}%[tb]
%\color{blue}
\label{alg:pseudo-code1}
\captionsetup{font=small}
\caption{PyTorch pseudocode for the enhanced IsoMax loss: Implementation.}
%\label{alg:MISA}
\definecolor{codeblue}{rgb}{0.25,0.5,0.5}
\lstset{
basicstyle=\fontsize{8.5 pt}{8.5 pt}\ttfamily\bfseries,
commentstyle=\fontsize{8.5 pt}{8.5 pt}\color{codeblue},
keywordstyle=\fontsize{8.5 pt}{8.5 pt}\color{magenta},
}
\begin{lstlisting}[language=python]
class IsoMaxPlusLossFirstPart(nn.Module):
    """This part replaces the model classifier output layer nn.Linear()"""
    def __init__(self, num_features, num_classes):
        super(IsoMaxPlusLossFirstPart, self).__init__()
        self.num_features = num_features
        self.num_classes = num_classes
        self.prototypes = nn.Parameter(torch.Tensor(num_classes, num_features))
        nn.init.normal_(self.prototypes, mean=0.0, std=1.0)
        self.distance_scale = nn.Parameter(torch.Tensor(1)) 
        nn.init.constant_(self.distance_scale, 1.0)

    def forward(self, features):
        distances = torch.abs(self.distance_scale) * torch.cdist(
            F.normalize(features), F.normalize(self.prototypes),
            p=2.0, compute_mode="donot_use_mm_for_euclid_dist")
        logits = -distances
        return logits

class IsoMaxPlusLossSecondPart(nn.Module):
    """This part replaces the nn.CrossEntropyLoss()"""
    def __init__(self, entropic_scale = 10.0):
        super(IsoMaxPlusLossSecondPart, self).__init__()
        self.entropic_scale = entropic_scale

    def forward(self, logits, targets):
        """Probabilities and logarithms are calculated separately and sequentially"""
        """Therefore, nn.CrossEntropyLoss() must not be used to calculate the loss"""
        distances = -logits
        probabilities_for_training = nn.Softmax(dim=1)(-self.entropic_scale * distances)
        probabilities_at_targets = probabilities_for_training[range(distances.size(0)), targets]
        loss = -torch.log(probabilities_at_targets).mean()
        return loss
\end{lstlisting}
\end{algorithm*}

Generalized ODIN \cite{Hsu2020GeneralizedOD} uses the \emph{in-distribution} validation set to avoid the need to access OOD samples to determine the hyperparameters required by the solution. However, considering that CIFAR10 and CIFAR100 do not have separate sets for validation and testing, the results presented in the paper may be overestimated because the validation sets used to define the hyperparameters were reused for OOD detection performance estimation. A more realistic OOD detection performance estimation could have been achieved by removing the in-distribution validation set from in-distribution training data, which would probably produce an even higher \emph{classification accuracy drop}. For example, \emph{Generalized ODIN produces a nearly 3\% classification accuracy drop} for ResNet34 trained on CIFAR100. Retraining the neural network using a very unusual fine-tuned regularization (dropout with $p\!\!=\!\!0.7$), as suggested by the authors, does not actually solve the problem \emph{because a classification accuracy drop of approximately 1\% persists and the OOD detection performance plunges approximately 10\%}.

The Generalized ODIN changes the architecture by adding a fully connected layer to the end of the network. We believe that this fact may explain the mentioned overfitting. We imagine that this problem may be even worse in datasets with fewer training examples. Additionally, the solution proposed in \cite{Hsu2020GeneralizedOD} is expensive and not environmentally friendly because it uses \emph{input preprocessing} and thus  produces slow and energy-inefficient inferences \cite{macdo2019isotropic, DBLP:journals/corr/abs-2006.04005}.

Recently, many OOD detection approaches have used additional/outlier/background data \cite{hendrycks2018deep, DBLP:journals/corr/abs-2010-03759, DBLP:conf/nips/DhamijaGB18}. In addition to requiring collecting additional data, these approaches require double the GPU memory size for training. The Gram matrices solution performs calculations on values produced by the model during inference \cite{Sastry2019DetectingOE} to perform OOD detection. This inference is efficient; however, the OOD detection has a high computational cost.

In some cases, an ensemble of classifiers is used \cite{vyas2018out}. For deep ensembles, \cite{lakshminarayanan2017simple} proposed an ensemble of same-architecture models trained with different random initial weights. %However, despite all the complexity and additional power processing required, this approach produces OOD detection performance lower than ODIN \cite{shafaei2018biased}.
Some proposals required structural changes in the model to perform OOD detection \cite{yu2019unsupervised}, and certain trials used uncertainty or confidence estimation/calibration techniques \cite{kendall2017uncertainties, Leibig2017LeveragingUI, malinin2018predictive, kuleshov2018accurate, subramanya2017confidence}. However, Bayesian neural networks that are used in most of these are typically more difficult to implement and require much more computational resources. Computational constraints also typically require approximations that compromise the performance, which is also affected by the prior distribution used \cite{lakshminarayanan2017simple}. For example, MC-dropout uses pretrained models with dropout activated during the test time. An average of many inferences is used to perform a single decision \cite{gal2016dropout}.

The \emph{entropic} out-of-distribution detection approach, which is composed of the IsoMax loss for training and the entropic score for OOD detection, avoids all mentioned special requirements and side effects \cite{macdo2019isotropic, DBLP:journals/corr/abs-2006.04005}. Indeed, no hyperparameter tuning is required because \emph{the entropic scale is a global constant kept equal to ten for all combinations of datasets and models}. Even if we call the entropic scale hyperparameter, the IsoMax loss does not involve hyperparameter \emph{tuning} because the same constant value of entropic scale is used in all cases. This result is possible because Mac\^{e}do et al. showed in \cite[Fig. 3]{macdo2019isotropic} and in \cite[Section IV, A]{DBLP:journals/corr/abs-2006.04005} that \emph{the OOD detection performance exhibits a well-behaved dependence on the entropic scale regardless of the dataset and model}. %In this paper, we also used the entropic scale equals to ten for all combinations of datasets and models.
No additional/outlier/background data are necessary. Models trained using IsoMax loss produce inferences as fast and energy-efficient as the inferences produced by softmax-loss-trained networks. OOD detection requires only a speedy entropy calculation. Finally, no classification accuracy drop is observed.

%%%%%%%%%%%%%%%%%%%%%
\textcolor{black}{For real-world large-scale applications, we require efficient inference and high-performance solutions. By \emph{efficient inference}, we mean solutions that do not increase the inference delay, computation, and energy consumption when compared with current softmax-loss-trained neural networks. By \emph{high performance}, we mean solutions that do not present a classification accuracy drop and \emph{simultaneously} exhibit much higher out-of-distribution detection when compared with usual SoftMax loss trained neural networks. In this paper, we are only interested in studying and comparing inference-efficient high-performance out-of-distribution detection solutions.}
%%%%%%%%%%%%%%%%%%%%%

Starting from the IsoMax loss, we added the following contributions. First, we normalize both the prototypes and the features. Second, we change the initialization of the prototypes. Third, we add the \emph{distance scale}, which is a \emph{learnable scalar parameter} that multiplies the \emph{feature-prototype distances}. We call the combination of these three modifications the \emph{isometrization} of the \emph{feature-prototype distances}. We call the proposed modified version of IsoMax the \emph{enhanced} isotropy maximization loss or the \emph{enhanced} IsoMax loss (IsoMax+ loss). Fourth, we use the \emph{minimum feature-prototype distance} as a score to perform OOD detection. Considering that the minimum feature-prototype distance is calculated to perform the classification, \emph{the OOD detection task presents essentially zero computational cost} because we simply reuse this value as a score to perform OOD detection. Fifth, in addition to experimental evidence, we provide insights into why a combination of training using IsoMax+ and performing OOD detection using the \emph{minimum distance score} produces a substantial performance increase in OOD detection.

Our approach keeps the solution seamless (i.e., avoids the previously mentioned special requirements and side effects) while significantly increasing the OOD detection performance. Similar to IsoMax loss, IsoMax+ works as a \emph{SoftMax loss drop-in replacement} because no procedures other than regular neural network training are required.

\section{Background}

Many approaches have been proposed to manage out-of-distribution detection. We may roughly classify them into three classes: training/fine-tuning methods and inference methods. Training/fine-tuning methods are used to train from scratch or to fine-tune pretrained models. Inference methods are applied to pretrained models (i.e., no training or fine-tuning is allowed), regardless of the training method used. Thus, training/fine-tuning and inference methods are complementary rather than competitors.

\subsection{Training/Fine-tuning Methods}

The first and most common training method is simply training with SoftMax loss and using the maximum probability score (MPS) \cite{hendrycks2017baseline} for OOD detection. Despite not presenting high OOD detection performance, this approach is seamless (i.e., no hyperparameters to tune, no additional data are necessary, efficient inference and OOD detection, no classification accuracy drop).

\begin{algorithm*}%[tb]
%\color{blue}
\label{alg:pseudo-code2}
\captionsetup{font=small}
\caption{PyTorch pseudocode for the enhanced IsoMax loss: Interface.}
%\label{alg:MISA}
\definecolor{codeblue}{rgb}{0.25,0.5,0.5}
\lstset{
basicstyle=\fontsize{8.5 pt}{8.5 pt}\ttfamily\bfseries,
commentstyle=\fontsize{8.5 pt}{8.5 pt}\color{codeblue},
keywordstyle=\fontsize{8.5 pt}{8.5 pt}\color{magenta},
}
\begin{lstlisting}[language=python]
class Model(nn.Module):
    def __init__(self):
    (...)
    #self.classifier = nn.Linear(num_features, num_classes)
    self.classifier = losses.IsoMaxPlusLossFirstPart(num_features, num_classes)
    
model = Model()
#criterion = nn.CrossEntropyLoss()
criterion = losses.IsoMaxPlusLossSecondPart()

outputs = model(inputs)
# outputs are equal to logits, which in turn are equivalent to negative distances
score = outputs.max(dim=1)[0] # this is the minimum distance score for detection
# the minimum distance score is the best option for the IsoMax+ loss
\end{lstlisting}
\end{algorithm*}

Techapanurak et al. proposed the Scaled Cosine approach \cite{techapanurak2019hyperparameterfree} by adding a layer to the output of the neural network to learn a scale to be multiplied by the cosine similarity that replaces the affine transformation in the SoftMax loss. The solution requires changing the training of the last layer by removing its weight decay to work properly. Therefore, this method does not work as a SoftMax loss drop-in replacement. Additionally, the solution is not seamless because \emph{it produces a classification accuracy drop higher than two percent in some cases}.

Hsu et al. developed the Generalized ODIN solution \cite{Hsu2020GeneralizedOD}. Despite having hyperparameters, this solution exclusively uses \emph{in-distribution} validation data to fine-tune the hyperparameters, preventing access to out-of-distribution data that is typically unavailable. Similar to ODIN, the solution uses \emph{input preprocessing}. The solution is not seamless because, despite preventing access to out-of-distribution data, it requires hyperparameter tuning. This solution also produces inefficient inferences and decreases classification accuracy in some cases.

Considering that we can train a model from scratch or fine-tune pretrained models to the proposed custom data, training methods are typically useful in practice. However, except for the pure SoftMax loss and the IsoMax loss variants, other training methods (e.g., Scaled Cosine, Generalized ODIN) decrease classification accuracy, probably due to the fact that those approaches add a fully connected layer to the end of the neural network, which makes it more likely to overfit. Additionally, in some cases, even inefficient inferences are produced \cite{Hsu2020GeneralizedOD}. To our knowledge, the pure SoftMax and IsoMax loss variants are the only seamless training methods currently available.

\subsubsection{Additional-Data Techniques}

Training/fine-tuning methods may be enhanced by additional-data techniques. The additional data are from an unlabeled third distribution different from both the in-distribution and the out-of-distribution. This fact is important because out-of-distribution data is often unknown in practice. The additional data are usually called outlier data or background samples in the literature \cite{hendrycks2018deep, DBLP:journals/corr/abs-2010-03759, DBLP:conf/nips/DhamijaGB18}.

\subsection{Inference Methods}

Liang et al. proposed ODIN \cite{liang2018enhancing}, which is essentially a combination of \emph{input preprocessing} and \emph{temperature calibration}. Input preprocessing consists of increasing the SoftMax score of any given input in a procedure that is inspired by adversarial attacks. Additionally, input preprocessing uses out-of-distribution samples to perform \emph{temperature calibration}.

Lee at al. proposed the Mahalanobis approach \cite{lee2018simple}, which consists of training an ad hoc generative classifier on the features extracted from intermediate layers of a pretrained model. OOD detection uses a regression model based on the Mahalanobis distances that are calculated using many layer activations (\emph{feature ensemble}). Additionally, this approach uses \emph{input preprocessing} to maximize performance. To avoid overestimating the OOD detection performance, this approach uses adversarial examples rather than out-of-distribution samples to validate hyperparameters because in real-world applications, we typically do not know out-of-distribution data.

Sastry et al. developed the Gram matrices method \cite{Sastry2019DetectingOE}, which identifies inconsistencies between activity patterns and predicted classes. This method characterizes activity patterns by Gram matrices and may be used to perform OOD detection.

First, we may believe that inference methods should be preferred due to avoiding training or fine-tuning models. However, in the real world, we typically have custom data that require us to train from scratch or at least fine-tune the proposed model. In such cases, we have no good reason not to train/fine-tune the model with a \emph{seamless training method} to start from a better baseline, regardless of planning to subsequently using an inference method. We emphasize that \emph{training and inference methods are complementary rather than competitors}.

Additionally, all known inference methods produce inference or require expensive additional computation to perform OOD detection. Thus, not using a training method yields higher computational costs and more energy-inefficient inferences during the thousands or millions of times the system will operate in the field. Regardless, inference methods may be applied after a training method to improve performance if the previously mentioned drawback is not an issue in a particular application.

\section{Enhanced IsoMax and the Distance Score}\label{sec:isomax2_loss}

In this section, we present the enhanced IsoMax for training and the minimum distance score for inference. By combining those methods, we develop a seamless, scalable, and high-performance out-of-distribution detection approach.

\subsection{Enhanced Isotropy Maximization Loss}

We consider an input $\bm{x}$ applied to a neural network that performs a parametrized transformation $\bm{f}_{\bm{\theta}}(\bm{x})$. We also consider $\bm{p}_{\bm{\phi}}^j$ to be the \emph{learnable prototype} associated with class $j$. Additionally, let the expression $\norm{\bm{f}_{\bm{\theta}}(\bm{x})\!-\!\bm{p}_{\bm{\phi}}^j}$ represent the \emph{nonsquared Euclidean distance} between $\bm{f}_{\bm{\theta}}(\bm{x})$ and $\bm{p}_{\bm{\phi}}^j$. Finally, we consider $\bm{p}_{\bm{\phi}}^k$ as the learnable prototype associated with the correct class for the input $\bm{x}$. Thus, we write the IsoMax loss \cite{macdo2019isotropic} using the equation below:

\begin{align}
\mathcal{L}_{\textsf{IsoMax}}&=-\log^\dagger\left(\frac{\exp(-E_s\norm{\bm{f}_{\bm{\theta}}(\bm{x})\!-\!\bm{p}_{\bm{\phi}}^k})}{\sum\limits_{j}\exp(-E_s\norm{\bm{f}_{\bm{\theta}}(\bm{x})\!-\!\bm{p}_{\bm{\phi}}^j})}\right)\label{eq:isomax_loss_full}
\end{align}
\blfootnote{\textsuperscript{$\dagger$}\emph{The probability (i.e., the expression between the outermost parentheses) and logarithm operations are computed sequentially and separately for higher OOD detection performance \cite{macdo2019isotropic} (see the source code).}}

%\begin{comment}
\begin{table*}%[h]
\centering
%\small
%\renewcommand{\arraystretch}{0.5}
\caption[caption]{\textbf{Classification accuracy of models trained using SoftMax, IsoMax, and IsoMax+ losses}.}
\vskip -0.25cm
\label{tab:classification_performance}
\begin{tabularx}{\textwidth}{clCCC}
%\begin{tabularx}{\columnwidth}{ll|CC}
\toprule
\multirow{2}{*}{\begin{tabular}[c]{@{}c@{}}Model\end{tabular}} & \multirow{2}{*}{\begin{tabular}[c]{@{}c@{}}Data\end{tabular}}
& Train Accuracy (\%) [$\uparrow$] & Test Accuracy (\%) [$\uparrow$]\\
&& \multicolumn{2}{c}{SoftMax Loss / IsoMax Loss / IsoMax+ Loss}\\
\midrule
\multirow{4}{*}{\begin{tabular}[c]{@{}c@{}}DenseNet100\end{tabular}} 
& CIFAR10 & 99.9$\pm$0.1 / 99.9$\pm$0.1 / 99.9$\pm$0.1 & 95.3$\pm$0.2 / 95.2$\pm$0.3 / 95.3$\pm$0.1\\
& CIFAR100 & 99.9$\pm$0.1 / 99.0$\pm$0.1 / 99.9$\pm$0.1 & 77.2$\pm$0.3 / 77.3$\pm$0.4 / 77.0$\pm$0.3\\
& SVHN & 97.0$\pm$0.2 / 97.8$\pm$0.2 / 97.0$\pm$0.3 & 96.5$\pm$0.2 / 96.6$\pm$0.3 / 96.5$\pm$0.2\\
\midrule
\multirow{3}{*}{\begin{tabular}[c]{@{}c@{}}ResNet110\end{tabular}} 
& CIFAR10 & 99.9$\pm$0.1 / 99.9$\pm$0.1 / 99.9$\pm$0.1 & 94.4$\pm$0.3 / 94.5$\pm$0.3 / 94.6$\pm$0.2\\
& CIFAR100 & 99.5$\pm$0.1 / 99.9$\pm$0.1 / 99.8$\pm$0.1 & 72.7$\pm$0.2 / 74.1$\pm$0.4 / 73.9$\pm$0.3\\
& SVHN & 99.8$\pm$0.1 / 99.9$\pm$0.1 / 99.5$\pm$0.1 & 96.6$\pm$0.3 / 96.8$\pm$0.2 / 96.9$\pm$0.3\\
\bottomrule
\end{tabularx}
\begin{justify}
Besides preserving classification accuracy compared with SoftMax loss- and IsoMax loss-trained networks, IsoMax+ loss-trained models show higher OOD detection performance. Results are shown as means and standard deviations of five different iterations (see Table~\ref{tbl:fair_odd}).
\end{justify}
\end{table*}
%\end{comment}

\begingroup
\begin{table*}%[!t]
%\renewcommand{\arraystretch}{1.25}
%\small
\centering
%\resizebox{\columnwidth}{!}{
\caption[caption]{\textbf{Fair comparison of seamless approaches: No hyperparameter tuning, no additional/outlier/background data,\\no classification accuracy drop, and no slow/inefficient inferences.} %LOSS+SCORE means training using the loss given by LOSS and performing OOD detection using the score given by SCORE.
}
\vskip -0.25cm
\begin{tabularx}{\textwidth}{lll|CC}
\toprule
%%%%%%%%%%%%%%%%%%%%%%%%%%%%%%%%%%%%%%%%%%%%%%%%%%%%%%%%%%%%%%%%%%%%%%%%%%%%%%%
\multirow{4}{*}{\begin{tabular}[c]{@{}c@{}}\\Model\end{tabular}} & \multirow{4}{*}{\begin{tabular}[c]{@{}c@{}}\\Data\\(training)\end{tabular}} & \multirow{4}{*}{\begin{tabular}[c]{@{}c@{}}\\OOD\\(unseen)\end{tabular}} & 
%%%%%%%%%%%%%%%%%%%%%%%%%%%%%%%%%%%%%%%%%%%%%%%%%%%%%%%%%%%%%%%%%%%%%%%%%%%%%%%%
%%%%%%%%%%%%%%%%%%%%%%%%%%%%%%%%%%%%%%%%%%%%%%%%%%%%%%%%%%%%%%%%%%%%%%%%%%%%%%%%%
\multicolumn{2}{c}{Out-of-Distribution Detection:}\\
%&&& \multicolumn{2}{c}{Accurate, Fast, Energy-Efficient, Scalable, Turnkey, and Seamless}\\
&&& \multicolumn{2}{c}{Seamless Approaches.}\\
\cmidrule{4-5}
&&& TNR@TPR95 (\%) [$\uparrow$] & AUROC (\%) [$\uparrow$] \\
&&& \multicolumn{2}{c}{SoftMax$_{\text{ES}}$ / IsoMax$_{\text{ES}}$ / IsoMax+$_{\text{MDS}}$ (ours)}\\
%%%%%%%%%%%%%%%%%%%%%%%%%%%%%%%%%%%%%%%%%%%%%%%%%%%%%%%%%%%%%%%%%%%%%%%%%%%%%%%%%
\midrule
\multirow{9}{*}{\begin{tabular}[c]{@{}c@{}}DenseNet100\end{tabular}}
& \multirow{3}{*}{\begin{tabular}[c]{@{}c@{}}CIFAR10\end{tabular}} 
& SVHN & 40.4$\pm$5.3 / 78.6$\pm$9.0 / \bf97.0$\pm$0.7 & 89.1$\pm$2.2 / 96.2$\pm$1.0 / \bf99.4$\pm$0.1\\
&& TinyImageNet & 58.0$\pm$3.7 / 83.9$\pm$3.6 / \bf92.6$\pm$2.4 & 94.0$\pm$0.6 / 97.1$\pm$0.4 / \bf98.5$\pm$0.3\\
&& LSUN & 64.6$\pm$1.7 / 90.3$\pm$1.4 / \bf94.3$\pm$1.4 & 95.2$\pm$0.4 / 98.0$\pm$0.3 / \bf99.1$\pm$0.2\\
\cmidrule{2-5} 
& \multirow{3}{*}{\begin{tabular}[c]{@{}c@{}}CIFAR100\end{tabular}} 
& SVHN & 21.9$\pm$2.8 / 29.6$\pm$3.7 / \bf78.2$\pm$4.1 & 78.4$\pm$3.3 / 88.8$\pm$2.8 / \bf96.3$\pm$1.3\\
&& TinyImageNet & 24.0$\pm$3.0 / 49.3$\pm$4.9 / \bf85.5$\pm$2.8 & 75.5$\pm$2.1 / 90.8$\pm$2.1 / \bf97.5$\pm$0.8\\
&& LSUN & 24.9$\pm$2.9 / 60.6$\pm$6.6 / \bf78.3$\pm$3.9 & 77.2$\pm$3.2 / 93.0$\pm$1.4 / \bf97.2$\pm$1.3\\
\cmidrule{2-5} 
& \multirow{3}{*}{\begin{tabular}[c]{@{}c@{}}SVHN\end{tabular}} 
& CIFAR10 & 85.2$\pm$3.3 / 93.4$\pm$1.1 / \bf95.1$\pm$0.4 & 97.3$\pm$0.3 / 98.4$\pm$0.2 / \bf99.1$\pm$0.1\\
&& TinyImageNet & 90.8$\pm$0.6 / 96.2$\pm$0.9 / \bf98.1$\pm$0.3 & 98.3$\pm$0.2 / 99.0$\pm$0.2 / \bf99.6$\pm$0.1\\
&& LSUN & 87.9$\pm$0.8 / 94.3$\pm$1.7 / \bf97.7$\pm$1.1 & 97.8$\pm$0.3 / 98.7$\pm$0.2 / \bf99.6$\pm$0.2\\
\midrule
\multirow{9}{*}{\begin{tabular}[c]{@{}c@{}}ResNet110\end{tabular}}
& \multirow{3}{*}{\begin{tabular}[c]{@{}c@{}}CIFAR10\end{tabular}} 
& SVHN & 35.6$\pm$5.0 / 68.7$\pm$5.5 / \bf85.3$\pm$3.9 & 88.9$\pm$0.9 / 94.8$\pm$1.5 / \bf98.1$\pm$0.9\\
&& TinyImageNet & 39.0$\pm$4.1 / 65.4$\pm$5.6 / \bf76.2$\pm$2.7 & 88.2$\pm$2.1 / 94.3$\pm$0.4 / \bf96.1$\pm$0.5\\
&& LSUN & 46.9$\pm$5.7 / 80.7$\pm$2.2 / \bf86.9$\pm$2.5 & 91.2$\pm$1.7 / 96.5$\pm$0.3 / \bf97.9$\pm$0.7\\
\cmidrule{2-5} 
& \multirow{3}{*}{\begin{tabular}[c]{@{}c@{}}CIFAR100\end{tabular}} 
& SVHN & 16.6$\pm$1.5 / 20.8$\pm$5.9 / \bf41.0$\pm$6.4 & 73.6$\pm$3.7 / 85.3$\pm$1.3 / \bf88.9$\pm$1.2\\
&& TinyImageNet & 15.6$\pm$2.1 / 22.8$\pm$1.9 / \bf44.7$\pm$5.9 & 71.5$\pm$1.7 / 81.3$\pm$2.3 / \bf88.0$\pm$2.9\\
&& LSUN & 16.5$\pm$3.3 / 22.9$\pm$3.5 / \bf46.1$\pm$4.3 & 72.6$\pm$4.2 / 83.3$\pm$2.0 / \bf89.1$\pm$2.4\\
\cmidrule{2-5} 
& \multirow{3}{*}{\begin{tabular}[c]{@{}c@{}}SVHN\end{tabular}} 
& CIFAR10 & 80.9$\pm$3.5 / 84.3$\pm$1.3 / \bf88.4$\pm$2.3 & 95.1$\pm$0.3 / 96.5$\pm$0.2 / \bf97.4$\pm$0.3\\
&& TinyImageNet & 84.0$\pm$3.2 / 87.4$\pm$2.2 / \bf93.4$\pm$1.8 & 96.6$\pm$0.7 / 96.7$\pm$0.6 / \bf98.5$\pm$0.7\\
&& LSUN & 81.4$\pm$2.3 / 84.7$\pm$2.1 / \bf90.2$\pm$2.2 & 96.1$\pm$0.3 / 95.8$\pm$0.4 / \bf97.8$\pm$0.3\\
\bottomrule
\end{tabularx}
\begin{justify}
SoftMax$_{\text{ES}}$ means training using SoftMax loss and performing OOD detection using the entropic score (ES). IsoMax$_{\text{ES}}$ means training using IsoMax loss and performing OOD detection using the entropic score (ES). IsoMax+$_{\text{MDS}}$ means training using IsoMax+ loss and performing OOD detection using minimum distance score (MDS). Results are shown as the mean and standard deviation of five iterations. The best results are shown in bold. See Table~\ref{tab:classification_performance}.% To our knowledge, IsoMax+$_{\text{ES}}$ presents \emph{state-or-the-art performance under these assumptions}.
\end{justify}
%}
\label{tbl:fair_odd}
\end{table*}
\endgroup

In the above equation, $E_s$ represents the entropic scale. From Equation~\eqref{eq:isomax_loss_full}, we observe that the distances from IsoMax loss are given by the expression $\mathcal{D}\!=\!\norm{\bm{f}_{\bm{\theta}}(\bm{x})\!-\!\bm{p}_{\bm{\phi}}^j}$. During inference, probabilities calculated based on these distances are used to produce the negative entropy, which serves as a score to perform OOD detection. However, because the features $\bm{f}_{\bm{\theta}}(\bm{x})$ are not normalized, examples with low norms are unjustifiably favored to be considered OOD examples because they tend to produce high entropy. Additionally, because the weights $\bm{p}_{\bm{\phi}}^j$ are not normalized, examples from classes that present prototypes with low norms are unjustifiably favored to be considered OOD examples for the same reason. Thus, we propose replacing $\bm{f}_{\bm{\theta}}(\bm{x})$ with its normalized version given by $\widehat{\bm{f}_{\bm{\theta}}(\bm{x})}\!=\!\bm{f}_{\bm{\theta}}(\bm{x})/\norm{\bm{f}_{\bm{\theta}}(\bm{x})}$. Additionally, we propose replacing $\bm{p}_{\bm{\phi}}^j$ with its normalized version given by $\widehat{\bm{p}_{\bm{\phi}}^j}\!=\!\bm{p}_{\bm{\phi}}^j/\norm{\bm{p}_{\bm{\phi}}^j}$. The expression $\norm{\bm{v}}$ represents the 2-norm of a given vector $\bm{v}$.

However, while the distances in the original IsoMax loss may vary from zero to infinity, the distance between two normalized vectors is always equal to or lower than two. To avoid this unjustifiable and unreasonable restriction, we introduce the \emph{distance scale} $d_s$, which is a \emph{scalar learnable parameter}. Naturally, we require the distance scale to always be positive by taking its absolute value $\abs{d_s}$.

Feature normalization makes the solution isometric regardless of the norm of the features produced by the examples. The distance scale is class independent because it is a \emph{single} scalar value learnable during training. The weight normalization and the class independence of the distance scale make the solution isometric regarding all classes. The final distance is isometric because it produces an isometric treatment of all features, prototypes, and classes.
Therefore, we can write the \emph{isometric distances} used by the IsoMax+ loss as $\mathcal{D_I}=\abs{d_s}\:\norm{\widehat{\bm{f}_{\bm{\theta}}(\bm{x})}\!-\!\widehat{\bm{p}_{\bm{\phi}}^j}}$.
%
%\begin{align}
%\mathcal{D_I}=\abs{d_s}\:\norm{\widehat{\bm{f}_{\bm{\theta}}(\bm{x})}\!-\!\widehat{\bm{p}_{\bm{\phi}}^j}}
%\label{eq:isomax2_logits}
%\end{align}
%
%Returning to Equation~\eqref{eq:isomax_loss_full}, we can write the expression for the IsoMax+ loss as follows (see Algorithm 1):
Returning to Equation~\eqref{eq:isomax_loss_full}, we can finally write the expression for the IsoMax+ loss as follows (see Algorithm 1):

\begin{equation}
\begin{aligned}
%\vskip -0.5cm
&\mathcal{L}_{\textsf{IsoMax+}}=\\
&-\log^{\dagger\dagger}\left(\frac{\exp(-E_s\:\abs{d_s}\:\norm{\widehat{\bm{f}_{\bm{\theta}}(\bm{x})}\!-\!\widehat{\bm{p}_{\bm{\phi}}^k}})}{\sum\limits_{j}\exp(-E_s\:\abs{d_s}\:\norm{\widehat{\bm{f}_{\bm{\theta}}(\bm{x})}\!-\!\widehat{\bm{p}_{\bm{\phi}}^j}})}\right)\label{eq:isomax2_loss_full}%\text{something\footnotemark}
\end{aligned}
\end{equation}
\blfootnote{\textsuperscript{$\dagger$}\textsuperscript{$\dagger$}\emph{Similar to \cite{macdo2019isotropic}, the probability (i.e., the expression between the outermost parentheses) and logarithm operations are computed sequentially and separately for higher OOD detection performance (see Algorithm 1 and the source code).}}

Applying the entropy maximization trick (i.e., the removal of the entropic scale $E_s$ for inference) \cite{macdo2019isotropic}, we can write the expression for the IsoMax+ loss probabilities used during inference for performing OOD detection when using the maximum probability score or the entropic score \cite{macdo2019isotropic}:

%\begin{comment}
%\begin{multline}\label{eq:probability_specific_diamax}
\begin{align}\label{eq:probability_class_isomax2}
\mathcal{P}_{\textsf{IsoMax+}}(y^{(i)}|\bm{x})&=\frac{\exp(-\:\abs{d_s}\:\norm{\widehat{\bm{f}_{\bm{\theta}}(\bm{x})}\!-\!\widehat{\bm{p}_{\bm{\phi}}^i}})}{\sum\limits_{j}\exp(-\:\abs{d_s}\:\norm{\widehat{\bm{f}_{\bm{\theta}}(\bm{x})}\!-\!\widehat{\bm{p}_{\bm{\phi}}^j}})}
\end{align}
%\end{multline}
%\end{comment}

\begingroup
\begin{table*}%[!t]
%\renewcommand{\arraystretch}{1.25}
%\small
\centering
%\resizebox{\columnwidth}{!}{
\caption[caption]{\textbf{Unfair comparison with approaches that use input preprocessing and produce slow/inefficient inferences\\in addition to requiring validation using adversarial examples.}}
\vskip -0.25cm
%\begin{tabularx}{\textwidth}{CCC|CC}
\begin{tabularx}{\textwidth}{lll|CC}
\toprule
%%%%%%%%%%%%%%%%%%%%%%%%%%%%%%%%%%%%%%%%%%%%%%%%%%%%%%%%%%%%%%%%%%%%%%%%%%%%%%%%%%
\multirow{4}{*}{\begin{tabular}[c]{@{}c@{}}\\Model\end{tabular}} & \multirow{4}{*}{\begin{tabular}[c]{@{}c@{}}\\Data\\(training)\end{tabular}} & \multirow{4}{*}{\begin{tabular}[c]{@{}c@{}}\\OOD\\(unseen)\end{tabular}} & 
%%%%%%%%%%%%%%%%%%%%%%%%%%%%%%%%%%%%%%%%%%%%%%%%%%%%%%%%%%%%%%%%%%%%%%%%%%%%%%%%%
\multicolumn{2}{c}{Comparison with approaches that use}\\
&&& \multicolumn{2}{c}{input preprocessing and adversarial validation.}\\
\cmidrule{4-5}
&&& AUROC (\%) [$\uparrow$] & DTACC (\%) [$\uparrow$] \\
&&& \multicolumn{2}{c}{ODIN / Mahalanobis / IsoMax+$_{\text{MDS}}$ (ours)}\\
%%%%%%%%%%%%%%%%%%%%%%%%%%%%%%%%%%%%%%%%%%%%%%%%%%%%%%%%%%%%%%%%%%%%%%%%%%%%%%%%%
\midrule
\multirow{6}{*}{\begin{tabular}[c]{@{}c@{}}DenseNet100\end{tabular}}
& \multirow{3}{*}{\begin{tabular}[c]{@{}c@{}}CIFAR10\end{tabular}} 
& SVHN & 92.1$\pm$0.2 / 97.2$\pm$0.3 / \bf99.4$\pm$0.1 & 86.1$\pm$0.3 / 91.9$\pm$0.3 / \bf96.3$\pm$0.4\\
&& TinyImageNet & 97.2$\pm$0.3 / 97.7$\pm$0.2 / \bf98.5$\pm$0.3 & 91.9$\pm$0.3 / {\bf94.3$\pm$0.5} / \bf93.9$\pm$0.6\\
&& LSUN & 98.5$\pm$0.3 / 98.6$\pm$0.2 / \bf99.1$\pm$0.2 & 94.3$\pm$0.3 / {\bf95.7$\pm$0.4} / \bf95.3$\pm$0.5\\
\cmidrule{2-5} 
& \multirow{3}{*}{\begin{tabular}[c]{@{}c@{}}CIFAR100\end{tabular}} 
& SVHN & 88.0$\pm$0.5 / 91.3$\pm$0.4 / \bf96.3$\pm$1.3 & 80.0$\pm$0.6 / 84.3$\pm$0.4 / \bf90.3$\pm$0.5\\
&& TinyImageNet & 85.6$\pm$0.5 / {\bf96.7$\pm$0.3} / \bf97.5$\pm$0.8 & 77.6$\pm$0.5 / {\bf91.0$\pm$0.4} / {\bf91.3$\pm$0.3}\\
&& LSUN & 85.7$\pm$0.6 / {\bf97.1$\pm$1.9} / {\bf97.2$\pm$1.3} & 77.5$\pm$0.4 / {\bf92.5$\pm$0.8} / \bf91.7$\pm$0.7\\
\midrule
\multirow{6}{*}{\begin{tabular}[c]{@{}c@{}}ResNet34\end{tabular}}
& \multirow{3}{*}{\begin{tabular}[c]{@{}c@{}}CIFAR10\end{tabular}} 
& SVHN & 86.0$\pm$0.3 / 95.0$\pm$0.3 / \bf98.0$\pm$0.4 & 77.1$\pm$0.4 / 88.7$\pm$0.3 / \bf93.5$\pm$0.4\\
&& TinyImageNet & 92.6$\pm$0.3 / {\bf98.3$\pm$0.4} / 95.3$\pm$0.3 & 86.5$\pm$0.5 / {\bf94.8$\pm$0.3} / 90.0$\pm$0.4\\
&& LSUN & 93.0$\pm$0.4 / {\bf98.8$\pm$0.3} / 96.3$\pm$0.4 & 86.3$\pm$0.4 / {\bf96.8$\pm$0.4} / 92.1$\pm$0.5\\
\cmidrule{2-5} 
& \multirow{3}{*}{\begin{tabular}[c]{@{}c@{}}CIFAR100\end{tabular}} 
& SVHN & 71.0$\pm$0.4 / 84.0$\pm$0.6 / \bf88.0$\pm$0.7 & 68.0$\pm$0.5 / 77.3$\pm$0.7 / \bf82.1$\pm$0.4\\
&& TinyImageNet & 83.1$\pm$0.3 / 87.3$\pm$0.5 / \bf90.5$\pm$0.4 & 76.2$\pm$0.4 / {\bf84.0$\pm$0.4} / \bf84.2$\pm$0.5\\
&& LSUN & 81.0$\pm$0.3 / 82.0$\pm$0.5 / \bf88.6$\pm$0.6 & 75.2$\pm$0.3 / 79.3$\pm$0.5 / \bf82.5$\pm$0.4\\
\bottomrule
\end{tabularx}
\begin{justify}
ODIN and Mahalanobis were applied to models trained using SoftMax loss. These approaches compute at least four times slower and less power efficient inferences \cite{macdo2019isotropic} because they use input preprocessing. Their hyperparameters were validated using adversarial examples. \textcolor{black}{Additionally, Mahalanobis requires feature extraction for training ad-hoc models to perform OOD detection. Finally, feature ensemble was also used.} IsoMax+$_{\text{MDS}}$ (ours) means training using IsoMax+ loss and performing OOD detection using minimum distance score (MDS). \textcolor{black}{No hyperparameter tuning is required when using IsoMax+ loss for training and the MDS for OOD detection.} \textcolor{black}{The IsoMax+ loss OOD detection performance shown in this table may be increased further without relying on input preprocessing or hyperparameter tuning by replacing the minimum distance score with the Mahalanobis \cite{lee2018simple} or the energy score~\cite{DBLP:journals/corr/abs-2010-03759}.} Results are the mean and standard deviation of five runs. The best results are shown in bold.
\end{justify}
%}
\label{tbl:unfair_odd}
\end{table*}
\endgroup

\begin{table*}%[t]
%\small
%\color{blue}
\renewcommand{\arraystretch}{1.25}
\centering
\caption{\textbf{Unfair comparison of outlier exposure-enhanced SoftMax loss with IsoMax loss and IsoMax+ loss without using additional data}.}
\vskip -0.25cm
\begin{tabularx}{\textwidth}{ll|CC}
\toprule
%%%%%%%%%%%%%%%%%%%%%%%%%%%%%%%%%%%%%%%%%%%%%%%%%%%%%%%%%%%%%%%%%%%%%%%%%%%%%%%%%%%%%%%
\multirow{4}{*}{\begin{tabular}[c]{@{}c@{}}\\Model\end{tabular}} & %\multirow{4}{*}{\begin{tabular}[c]{@{}c@{}}\\In-Data\\(training)\end{tabular}} & \multirow{4}{*}{\begin{tabular}[c]{@{}c@{}}\\OOD Detection\\Approach\end{tabular}} & 
\multirow{4}{*}{\begin{tabular}[c]{@{}c@{}}\\Data\\(training)\end{tabular}} & 
%%%%%%%%%%%%%%%%%%%%%%%%%%%%%%%%%%%%%%%%%%%%%%%%%%%%%%%%%%%%%%%%%%%%%%%%%%%%%%%%%%%%%%%
\multicolumn{2}{c}{Comparison of IsoMax loss variants without using additional data}\\
&& \multicolumn{2}{c}{with outlier exposure-enhanced SoftMax loss.}\\
\cmidrule{3-4}
&& TNR@TPR95 (\%) [$\uparrow$] & AUROC (\%) [$\uparrow$]\\
%&& SoftMax+OE / IsoMax / IsoMax+ & SoftMax+OE / IsoMax / IsoMax+\\
&& \multicolumn{2}{c}{SoftMax$^{\text{OE}}_{\text{ES}}$ / IsoMax$_{\text{ES}}$ / IsoMax+$_{\text{MDS}}$ (ours)}\\
%%%%%%%%%%%%%%%%%%%%%%%%%%%%%%%%%%%%%%%%%%%%%%%%%%%%%%%%%%%%%%%%%%%%%%%%%%%%%%%%%%%%%%%
\midrule
\multirow{2}{*}{\begin{tabular}[c]{@{}c@{}}DenseNet100\end{tabular}}
& \multirow{1}{*}{\begin{tabular}[c]{@{}c@{}}CIFAR10\end{tabular}} 
& {\bf94.4$\pm$1.4} / 84.2$\pm$4.6 / \bf94.6$\pm$1.5 & 98.0$\pm$0.3 / 97.1$\pm$0.5 / \bf99.0$\pm$0.2\\
& \multirow{1}{*}{\begin{tabular}[c]{@{}c@{}}CIFAR100\end{tabular}} 
& 36.4$\pm$9.4 / 46.5$\pm$5.0 / \bf80.6$\pm$3.6 & 83.8$\pm$5.6 / 90.8$\pm$2.1 / \bf97.0$\pm$1.1\\
%%& \multirow{1}{*}{\begin{tabular}[c]{@{}c@{}}SVHN\end{tabular}} 
%%& {\bf99.9} / 95.9 / 97.1 & {\bf99.9} / 98.9 / 99.3\\
\midrule
\multirow{2}{*}{\begin{tabular}[c]{@{}c@{}}ResNet110\end{tabular}}
& \multirow{1}{*}{\begin{tabular}[c]{@{}c@{}}CIFAR10\end{tabular}} 
& {\bf82.4$\pm$2.1} / 71.6$\pm$4.4 / \bf82.8$\pm$3.0 & {\bf96.8$\pm$0.7} / 95.2$\pm$0.7 / \bf97.3$\pm$0.7\\
& \multirow{1}{*}{\begin{tabular}[c]{@{}c@{}}CIFAR100\end{tabular}} 
& 29.1$\pm$4.5 / 22.1$\pm$3.7 / \bf43.9$\pm$5.5 & 80.5$\pm$2.8 / 83.3$\pm$1.8 / \bf88.6$\pm$2.1\\
%%& \multirow{1}{*}{\begin{tabular}[c]{@{}c@{}}SVHN\end{tabular}} 
%%& {\bf99.9} / 81.7 / 77.1 & {\bf99.9} / 95.1 / 94.6\\
\bottomrule
\end{tabularx}
\begin{justify}
SoftMax$^{\text{OE}}_{\text{ES}}$ means training using SoftMax loss enhanced during training using outlier exposure \cite{hendrycks2018deep}, which requires the collection of outlier data, and performing OOD detection using the entropic score. We used the same outlier data used in \cite{hendrycks2018deep}. We collected the same amount of outlier data as the number of training examples present in the training set used to train SoftMax$^{\text{OE}}$. Despite being possible \cite{DBLP:journals/corr/abs-2006.04005}, the IsoMax loss and IsoMax+ loss were not enhanced by outlier exposure to keep the solution seamless. IsoMax$_{\text{ES}}$ means training using IsoMax loss and performing OOD detection using the entropic score. IsoMax+$_{\text{MDS}}$ (ours) means training using IsoMax+ loss and performing OOD detection using minimum distance score (MDS). The values of the performance metrics TNR@TPR95 and AUROC were averaged over all out-of-distribution (SVHN, TinyImageNet, and LSUN). Results are shown as the mean and standard deviation of five iterations. The best results are shown in bold.
\end{justify}
\label{tbl:comp_oe}
\end{table*}

\begingroup
\begin{table*}%[t]
%\scriptsize
%\small
%\color{blue}
%\renewcommand{\arraystretch}{1.25}
\centering
\caption{\textbf{Comparison of IsoMax variants using different scores}.}
\vskip -0.25cm
\begin{tabularx}{\textwidth}{ll|CC}
\toprule
%%%%%%%%%%%%%%%%%%%%%%%%%%%%%%%%%%%%%%%%%%%%%%%%%%%%%%%%%%%%%%%%%%%%%%%%%%%%%%%%%%%%%%%
\multirow{4}{*}{\begin{tabular}[c]{@{}c@{}}\\Model\end{tabular}} & %\multirow{4}{*}{\begin{tabular}[c]{@{}c@{}}\\In-Data\\(training)\end{tabular}} & \multirow{4}{*}{\begin{tabular}[c]{@{}c@{}}\\OOD Detection\\Approach\end{tabular}} & 
\multirow{4}{*}{\begin{tabular}[c]{@{}c@{}}\\Data\\(training)\end{tabular}} & 
%%%%%%%%%%%%%%%%%%%%%%%%%%%%%%%%%%%%%%%%%%%%%%%%%%%%%%%%%%%%%%%%%%%%%%%%%%%%%%%%%%%%%%%
\multicolumn{2}{c}{Comparison of IsoMax loss variants}\\
&& \multicolumn{2}{c}{using diferent scores.}\\
\cmidrule{3-4}
&& TNR@TPR95 (\%) [$\uparrow$] & AUROC (\%) [$\uparrow$]\\
%&& SoftMax+OE / IsoMax / IsoMax+ & SoftMax+OE / IsoMax / IsoMax+\\
&& \multicolumn{2}{c}{IsoMax$_{\text{ES}}$ / IsoMax$_{\text{MDS}}$ / IsoMax+$_{\text{ES}}$ / IsoMax+$_{\text{MDS}}$ (ours)}\\
%%%%%%%%%%%%%%%%%%%%%%%%%%%%%%%%%%%%%%%%%%%%%%%%%%%%%%%%%%%%%%%%%%%%%%%%%%%%%%%%%%%%%%%
\midrule
\multirow{2}{*}{\begin{tabular}[c]{@{}c@{}}DenseNet100\end{tabular}}
& \multirow{1}{*}{\begin{tabular}[c]{@{}c@{}}CIFAR10\end{tabular}} 
& 84.2$\pm$4.6 / 0.9$\pm$0.5 / 89.3$\pm$2.3 / \bf94.6$\pm$1.5 & 97.1$\pm$0.5 / 65.5$\pm$6.0 / 97.9$\pm$0.3 / \bf99.0$\pm$0.2\\
& \multirow{1}{*}{\begin{tabular}[c]{@{}c@{}}CIFAR100\end{tabular}} 
& 46.5$\pm$5.0 / 6.2$\pm$6.1 / 63.7$\pm$8.0 / \bf80.6$\pm$3.6 & 90.8$\pm$2.1 / 50.1$\pm$7.4 / 94.0$\pm$1.3 / \bf97.0$\pm$1.1\\
%%& \multirow{1}{*}{\begin{tabular}[c]{@{}c@{}}SVHN\end{tabular}} 
%%& {\bf99.9} / 95.9 / 97.1 & {\bf99.9} / 98.9 / 99.3\\
\midrule
\multirow{2}{*}{\begin{tabular}[c]{@{}c@{}}ResNet110\end{tabular}}
& \multirow{1}{*}{\begin{tabular}[c]{@{}c@{}}CIFAR10\end{tabular}} 
& 71.6$\pm$4.4 / 0.1$\pm$0.1 / 74.8$\pm$3.5 / \bf82.8$\pm$3.0 & 95.2$\pm$0.7 / 83.7$\pm$4.1 / 95.2$\pm$0.6 / \bf97.3$\pm$0.7\\
& \multirow{1}{*}{\begin{tabular}[c]{@{}c@{}}CIFAR100\end{tabular}} 
& 22.1$\pm$3.7 / 1.6$\pm$2.0 / 22.3$\pm$5.3 / \bf43.9$\pm$5.5 & 83.3$\pm$1.8 / 61.9$\pm$7.2 / 84.4$\pm$0.8 / \bf88.6$\pm$2.1\\
%%& \multirow{1}{*}{\begin{tabular}[c]{@{}c@{}}SVHN\end{tabular}} 
%%& {\bf99.9} / 81.7 / 77.1 & {\bf99.9} / 95.1 / 94.6\\
\bottomrule
\end{tabularx}
\begin{justify}
IsoMax$_{\text{ES}}$ means training using IsoMax loss and performing OOD detection using the entropic score (ES). IsoMax$_{\text{MDS}}$ means training using IsoMax loss and performing OOD detection using the minimum distance score (MDS). IsoMax+$_{\text{ES}}$ means training using IsoMax+ loss and performing OOD detection using entropic score (ES). IsoMax+$_{\text{MDS}}$ (ours) means training using IsoMax+ loss and performing OOD detection using minimum distance score (MDS). The values of the performance metrics TNR@TPR95 and AUROC were averaged over all out-of-distribution (SVHN, TinyImageNet, and LSUN). Results are shown as the mean and standard deviation of five iterations. The best results are shown in bold. See Fig.~\ref{fig:histograms}.
\end{justify}
\label{tbl:diff_scores}
\end{table*}
\endgroup

Different from IsoMax loss, where the prototypes are initialized to a zero vector, we initialized all prototypes using a normal distribution with a mean of zero and a standard deviation of one. This approach is necessary because we normalize the prototypes when using IsoMax+ loss. The distance scale is initialized to one, and we added no hyperparameters to the solution.

\subsection{Minimum Distance Score}

Motivated by the desired characteristics of the isometric distances used in IsoMax+, we use the minimum distance as score for performing OOD detection. Naturally, the \emph{minimum distance score} for IsoMax+ is given by:

\begin{align}
%\mathcal{S}_{\textsf{MinDist}}\!=\!\min_j
\text{MDS}\!=\!\min_j \left(\norm{\widehat{\bm{f}_{\bm{\theta}}(\bm{x})}\!-\!\widehat{\bm{p}_{\bm{\phi}}^j}}\right)
%\:\abs{d_s}\:
\label{eq:min_distance_score}
\end{align}

\begin{figure*}%[h]
\centering
\subfloat[]{\includegraphics[width=0.9\textwidth]{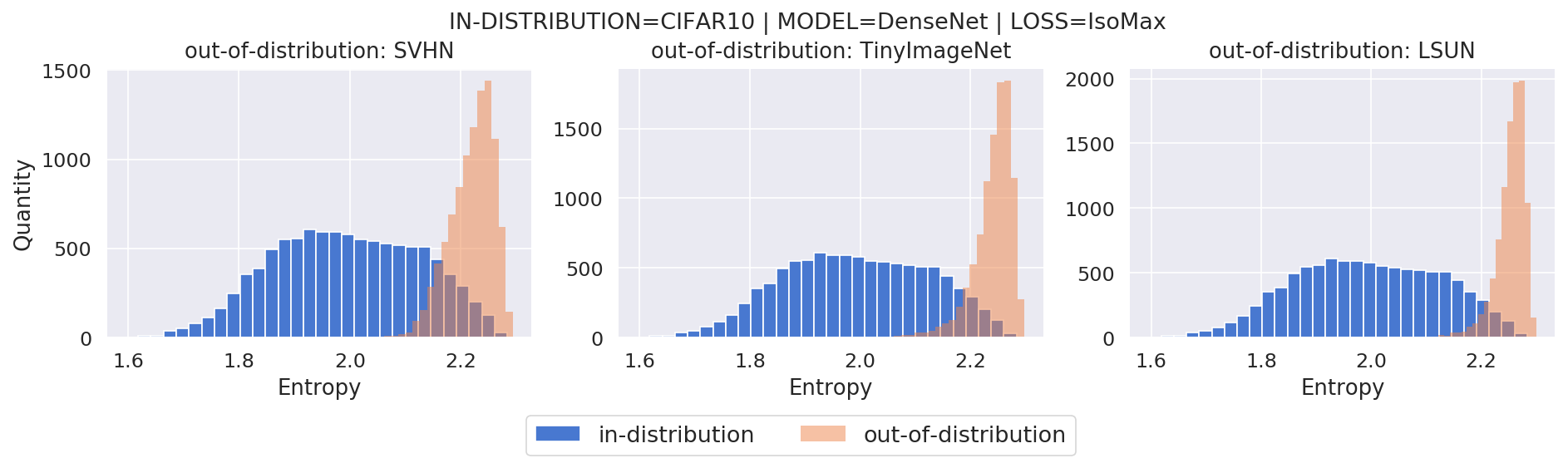}}
\\
\subfloat[]{\includegraphics[width=0.9\textwidth]{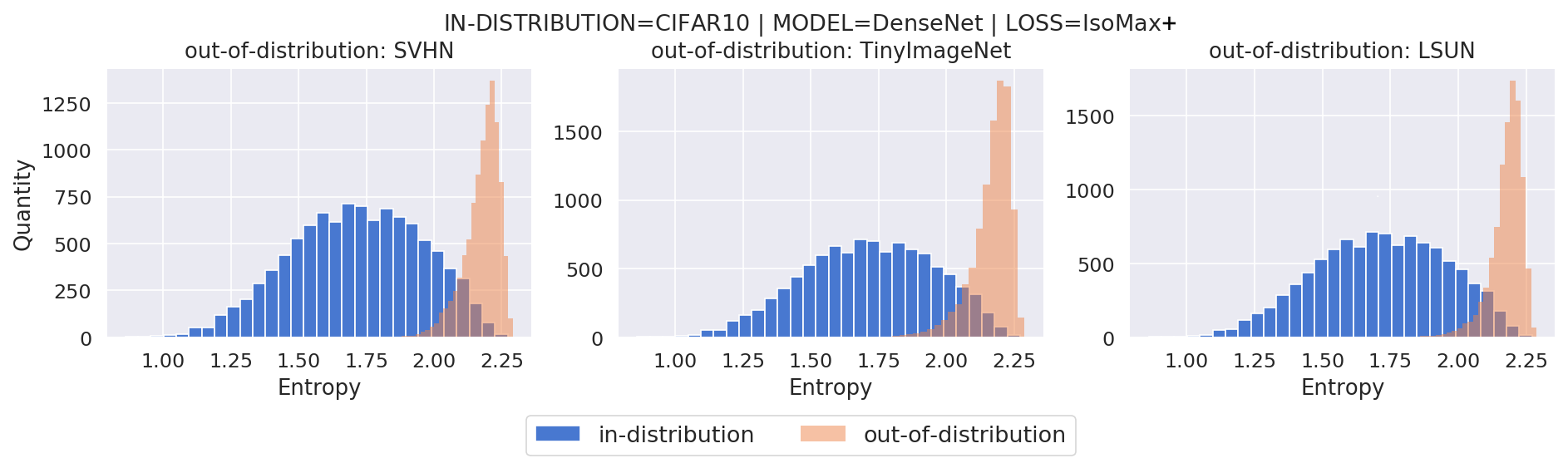}}
\\
\subfloat[]{\includegraphics[width=0.9\textwidth]{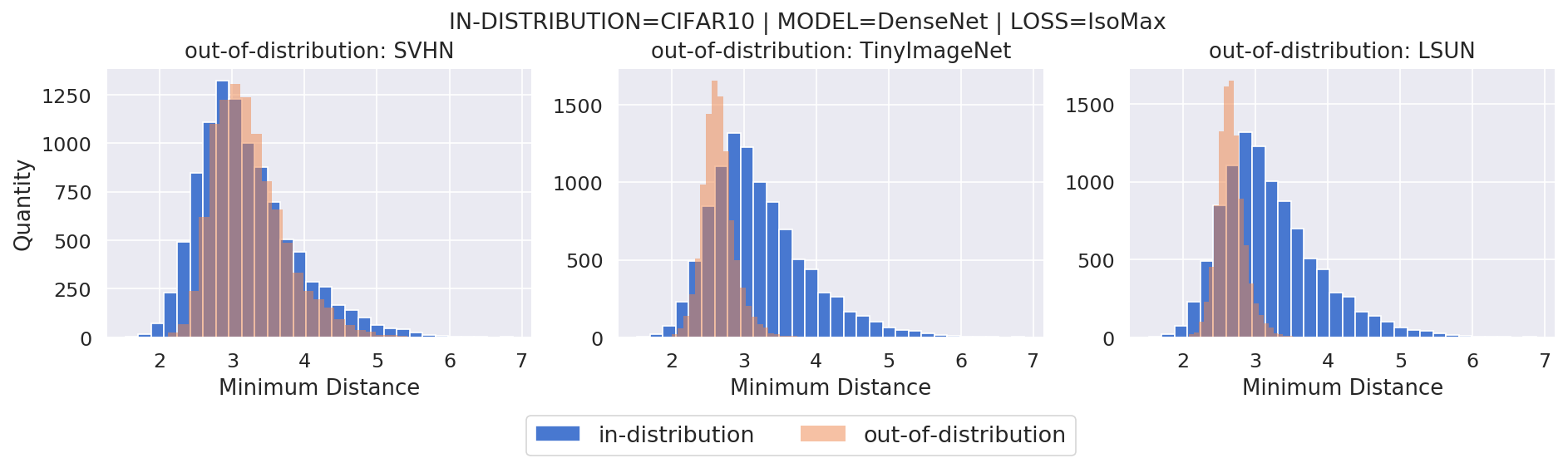}}
\\
\subfloat[]{\includegraphics[width=0.9\textwidth]{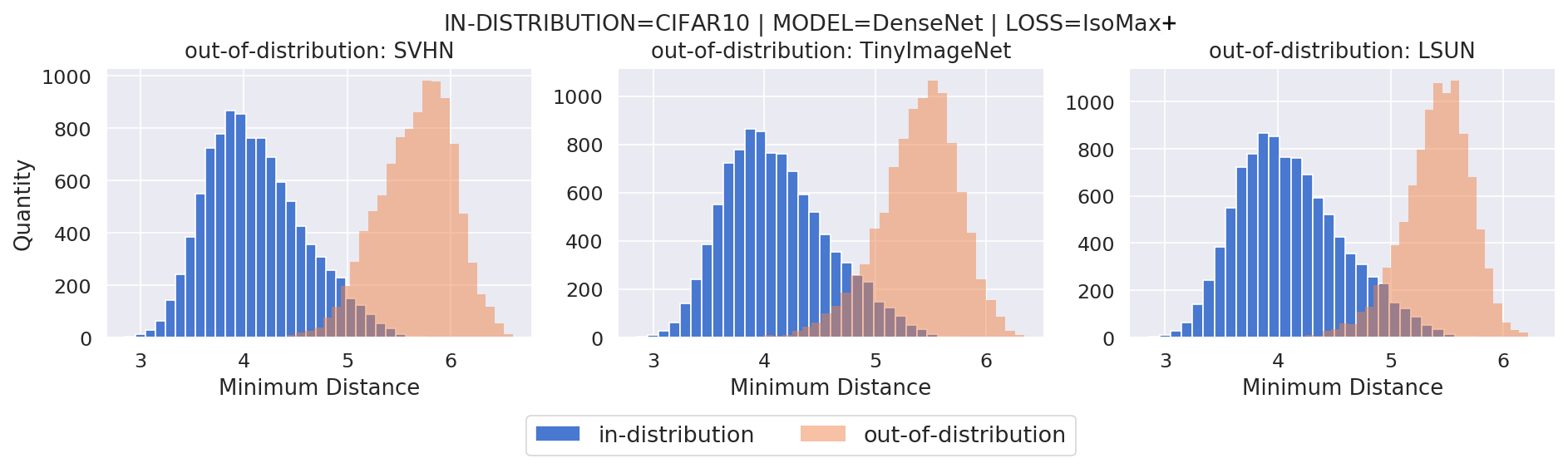}}
\\
\caption{In agreement with other studies \cite{DBLP:conf/nips/LiuLPTBL20}, (a) IsoMax (b) and IsoMax+ produce higher entropy/uncertainty on out-of-distributions than in-distribution. Therefore, the entropic score produces high OOD detection performance in both cases. (c) However, IsoMax does not make the in-distribution closer to the prototypes than out-of-distributions. (d) Concurrently, by introducing \emph{distance isometrization}, IsoMax+ gets in-distribution closer to the prototypes while pushing out-of-distribution data far away, which is what we expect based on the findings of other recent studies \cite{DBLP:conf/nips/LiuLPTBL20}. This result also explains why the minimum distance score produces high performance when using IsoMax+,  while producing low performance when using IsoMax. See also Table~\ref{tbl:diff_scores}.}
\label{fig:histograms}
\end{figure*}

In this equation, $\abs{d_s}$ was removed because, after training, it is a scale factor that does not affect the detection decision. Considering that the minimum distance is computed to perform the classification because the predicted class is the one that presents \emph{the lowest feature-prototype distance}, when using the minimum distance score, \emph{the OOD detection exhibits essentially zero latency and computational cost} because we simply reuse the minimum distance that was already calculated for classification purpose.

\section{Experiments}\label{sec:experiments}

To allow standardized comparison, we used the datasets, training procedures, and metrics that were established in \cite{hendrycks2017baseline} and used in many subsequent OOD detection papers \cite{liang2018enhancing, lee2018simple, Hein2018WhyRN}. We trained many 100-layer DenseNetBCs with growth rate $k\!=\!12$ (i.e., 0.8M parameters) \cite{Huang2017DenselyNetworks}, 110-layer ResNets (\emph{correct size proper implementation}) \cite{He_2016}\footnote{\url{https://github.com/akamaster/pytorch_resnet_cifar10}}, and 34-layer ResNets (\emph{overparametrized commonly used implementation}) \cite{He_2016}\footnote{\url{https://github.com/pokaxpoka/deep_Mahalanobis_detector}} on CIFAR10 \cite{Krizhevsky2009LearningImages}, CIFAR100 \cite{Krizhevsky2009LearningImages}, SVHN \cite{Netzer2011ReadingLearning}, and TinyImageNet \cite{Deng2009ImageNetDatabase} datasets with SoftMax, IsoMax, and IsoMax+ losses using the same procedures (e.g., initial learning rate, learning rate schedule, weight decay).

We used SGD with the Nesterov moment equal to 0.9 with a batch size of 64 and an initial learning rate of 0.1. The weight decay was 0.0001, and we did not use dropout. We trained during 300 epochs for CIFAR10, CIFAR100, and SVHN; and during 90 epochs for TinyImageNet. We used a learning rate decay rate equal to ten applied in epoch numbers 150, 200, and 250 for CIFAR10, CIFAR100, and SVHN; and 30 and 60 for TinyImageNet. The code to reproduce the results and replace the SoftMax loss by the IsoMax+ loss is available at \url{https://github.com/dlmacedo/entropic-out-of-distribution-detection}.

We used resized images from the TinyImageNet \cite{Deng2009ImageNetDatabase}, the Large-scale Scene UNderstanding dataset (LSUN) \cite{Yu2015LSUNLoop}, CIFAR10, and SVHN \cite{Netzer2011ReadingLearning} to create out-of-distribution samples. We added these out-of-distribution images to the validation sets of in-distribution data to form the test sets and evaluate the OOD detection performance.

%Uniform Noise: The uniform noise dataset is generated by drawing each pixel in a 32x32 RGB image from an i.i.d uniform distribution of the range [0, 1]. The dataset contains 10,000 samples in total. Gaussian Noise: The Gaussian noise dataset is generated by drawing each pixel in a 32x32 RGB image from an i.i.d Gaussian distribution with a mean of 0.5 and variance of 1. The pixel values in each image are clipped to the range [0,1] to keep them in the expected range for images. The dataset contains 10,000 samples in total.

We evaluated the OOD detection performance using the true negative rate at 95\% true positive rate (TNR@TPR95), the area under the receiver operating characteristic curve (AUROC) and the detection accuracy (DTACC), which corresponds to the maximum classification probability over all possible thresholds $\delta$:
\begin{align*}
1- \min_{\delta} \big\{ P_{\texttt{in}} \left( o \left( \mathbf{x} \right) \leq \delta \right) P \left(\mathbf{x}\text{ is from }P_{\texttt{in}}\right)\\+ P_{\texttt{out}} \left( o \left( \mathbf{x} \right) >\delta \right) P \left(\mathbf{x}\text{ is from }P_{\texttt{out}}\right)\big\},
\end{align*}
where $o(\mathbf{x})$ is the OOD detection score. It is assumed that both positive and negative samples have an equal probability of being in the test set, i.e., $P \left(\mathbf{x}\text{ is from }P_{\texttt{in}}\right) = P \left(\mathbf{x}\text{ is from }P_{\texttt{out}}\right)$. All metrics follow the calculation procedures specified in \cite{lee2018simple}.

In this paper, we did not directly present a comparison with the approaches that produce \emph{classification accuracy drop} (e.g., \cite{techapanurak2019hyperparameterfree, Hsu2020GeneralizedOD}) because this is a substantial limitation from a practical perspective \cite{carlini2019evaluating}. %Additionally, we do not compare to approaches that present inefficient inferences such as those that use \emph{input preprocessing} because this is a severe drawback from a cost and environment perspective. We believe that approaches that produce \emph{classification accuracy drop} or \emph{inefficient inferences} are unlikely to be deployed in large-scale projects in the real world.
Regardless, it is easy to notice that IsoMax+ presents a similar OOD detection performance of Generalized ODIN (difference lower than the margin of error alternatively in favor of one or another) while simultaneously avoiding classification accuracy drop, hyperparameter tuning, and inefficient inferences. Moreover, IsoMax+ is also much easier to implement and to use.

For example, on the one hand, when using Generalized ODIN, we need to exclude some components of the solution from receiving weight decay, which requires optimization procedure modifications. On the other hand, just a loss replacement is needed in our proposal. Additionally, Generalized ODIN has many flavors. Consequently, it is hard to know a prior which one will perform best for a given dataset and model without training all of them from scratch using the same data and model.%In this paper, no validation using OOD samples are considered.%As we are studying the problem from a OOD detection perspective rather than a uncertainty estimation one, we use metrics such as AUROC and detection accuracy rather than negative log likelihood (NLL) and Expected Calibration Error (ECE) \cite{Guo2017OnCO} or Brier score \cite{Brier1950VERIFICATIONOF}.

\section{Results and Discussion}\label{sec:results_discussion}

In this section, we present the results and discussion. We initially show that the enhanced IsoMax produces classification accuracies that are comparable to those of the SoftMax loss function. We then show that the enhanced IsoMax produces much higher OOD detection performance than the IsoMax Loss and the SoftMax Loss.

%\subsection{Classification Accuracy}

Table~\ref{tab:classification_performance} shows classification accuracies. We see that IsoMax+ loss never exhibits \emph{classification accuracy drop} compared to SoftMax loss or IsoMax loss regardless of the dataset and model. The IsoMax loss variants achieve more than one percent better accuracy than the SoftMax loss when using ResNet110 on the CIFAR100 dataset.

\subsection{Out-of-Distribution Detection}

Table~\ref{tbl:fair_odd} summarizes the results of the \emph{fair} OOD detection comparison. We report the results using the entropic score for SoftMax loss (SoftMax$_{\text{ES}}$) and IsoMax loss (IsoMax$_{\text{ES}}$) because this score always overcame the maximum probability score in these cases. For IsoMax+, we report the values using the minimum distance score (IsoMax+$_{\text{MDS}}$) because this method overcame the maximum probability and the entropic score in this situation.

All approaches are accurate (i.e., exhibit no \emph{classification accuracy drop}); fast and power-efficient (i.e., inferences are performed without \emph{input preprocessing}); and no hyperparameter tuning was performed. Additionally, no additional/outlier/background data are required. IsoMax+$_{\text{MDS}}$ always overcomes IsoMax$_{\text{ES}}$ performance, regardless of the model, dataset, and out-of-distribution data.

The minimum distance score produces high OOD detection performance when combined with IsoMax+, which shows that the isometrization of the distances works appropriately in this case. However, the same minimum distance score produced low OOD detection performance when combined with the original IsoMax loss. Fig.~\ref{fig:histograms}, and Table~\ref{tbl:diff_scores} provide an explanation for this fact.

Table~\ref{tbl:unfair_odd} summarizes the results of an \emph{unfair} OOD detection comparison because the methods have different requirements and produce distinct side effects. ODIN \cite{liang2018enhancing} and the Mahalanobis\footnote{Considering that the proposed approach easily outperforms the vanilla Mahalanobis method when applied to SoftMax loss trained models (i.e., training a Mahalanobis distance-based classifier using features extracted from the neural network and using the Mahalanobis distance as score), throughout this paper, we use the term Mahalanobis approach to refer to the full Mahalanobis approach.} \cite{lee2018simple} approaches require adversarial samples to be generated to validate hyperparameters for each combination of dataset and model. These approaches also use \emph{input preprocessing}, \emph{which makes inferences at least four times slower and less energy-efficient} \cite{macdo2019isotropic, DBLP:journals/corr/abs-2006.04005}. Validation using adversarial examples may be cumbersome when performed from scratch on novel datasets because hyperparameters such as optimal adversarial perturbations may be unknown in such cases. IsoMax+$_{\text{MDS}}$ does not have these requirements, and does not produce the side effects.

Additionally, IsoMax+$_{\text{MDS}}$ achieves higher performance than ODIN by a large margin. In addition to the differences between the entropy maximization trick and temperature calibrations present in \cite{macdo2019isotropic, DBLP:journals/corr/abs-2006.04005}, we emphasize that training with entropic scale affects the learning of all weights, while changing the temperature during inference affects only the last layer. Thus, the fact that the proposed solution overcomes ODIN by a safe margin is evidence that the \emph{entropy maximization trick produces much higher OOD detection performance than temperature calibration}, even when the latter is combined with input preprocessing. Moreover, the entropy maximization trick does not require access to validation data to tune the temperature.

In addition to being seamless and avoiding the drawbacks of the Mahalanobis approach, IsoMax+$_{\text{MDS}}$ typically overcomes it in terms of AUROC and produces similar performance when considering the DTACC. We emphasize that the IsoMax+ loss OOD detection performance presented in Table~\ref{tbl:unfair_odd} may increase further without requiring input preprocessing or hyperparameter tuning by replacing the minimum distance score with the Mahalanobis score \cite{lee2018simple}.

Table~\ref{tbl:comp_oe} \emph{unfairly} compares the performance of the proposed approach with the outlier exposure solution. Similar to IsoMax variants, the outlier exposure approach does not require hyperparameter tuning and produces efficient inferences. However, it does require collecting outlier data, while our approach does not. We emphasize that outlier exposure may also be combined with IsoMax loss variants to further increase the OOD detection performance \cite{DBLP:journals/corr/abs-2006.04005}. In the table, we present the IsoMax loss variants \emph{without outlier exposure} to show that the outlier exposure-enhanced SoftMax loss typically achieves worse OOD detection than IsoMax+$_{\text{MDS}}$ \emph{even without using outlier exposure}.

\begin{comment}
\subsection{Loss Landscapes}

\begin{figure*}%[h]
\centering
%\includegraphics[width=0.95\textwidth]{plots/plot_histogram_entropies+densenetbc100+cifar10+isomax_no_no_no_final.png}
\subfloat[]{\includegraphics[height=0.3\textwidth]{plots/softmax.png}}
\hspace{0.05cm}
\subfloat[]{\includegraphics[height=0.3\textwidth]{plots/isomax.png}}
\hspace{0.05cm}
\subfloat[]{\includegraphics[height=0.3\textwidth]{plots/isomaxplus.png}}
\\
\subfloat[]{\includegraphics[width=0.33\textwidth]{plots/softmax.pdf}}
%\hspace{0.05cm}
\subfloat[]{\includegraphics[width=0.33\textwidth]{plots/isomax.pdf}}
%\hspace{0.05cm}
\subfloat[]{\includegraphics[width=0.33\textwidth]{plots/isomaxplus.pdf}}
\caption{3D loss surfaces and 2D loss contours as proposed in \cite{DBLP:conf/nips/Li0TSG18}. (a) SoftMax, (b) IsoMax, and (c) IsoMaxPlus.}
\label{fig:loss_landscapes}
\end{figure*}

Fig. \ref{fig:loss_landscapes} shows the  3D surfaces and 2D contours of the losses. Considering that IsoMax outperforms SoftMax and IsoMax outperforms IsoMax, we concluded that less steep 3D inclination provides increased out-of-distribution detection performances. In other words, less concentration of 2D contours produces improved OOD detection performance. 
\end{comment}

\section{Conclusion}

In this paper, we improved the IsoMax loss function 
by making many important modifications to it. Additionally, we proposed the \emph{zero computational cost} minimum distance score. Experiments showed that these modifications achieve higher OOD detection performance while maintaining the desired benefits of IsoMax loss (i.e., absence of hyperparameters to tune, no reliance on additional/outlier/background data, fast and power-efficient inference, and no \emph{classification accuracy drop}).

Similar to IsoMax loss, after training using the proposed IsoMax+ loss, we may apply inference-based approaches (e.g., ODIN, Mahalanobis, Gram matrices, outlier exposure, energy-based) to the pretrained model to eventually increase the overall OOD detection performance even more. Thus, the IsoMax+ loss is a replacement for SoftMax loss but not for OOD methods that may be applied to pretrained models, which may be used to improve even more the OOD detection performance of IsoMax+ loss pretrained~networks.

There is no drawback in training a model using IsoMax+ loss instead of SoftMax loss or IsoMax loss, regardless of planning to subsequently use an inference-based OOD detection approach to further increase OOD detection performance. Therefore, instead of competitors, the OOD detection approaches that may be applied to pretrained models are actually complementary to the proposed approach \cite{macdo2019isotropic, DBLP:journals/corr/abs-2006.04005}. Hence, IsoMax+ loss constitutes a better baseline than SoftMax loss or IsoMax loss to construct future OOD detection methods.

In future work, we plan to verify whether our approach works satisfactorily when dealing with text and audio datasets. We also intend to verify the performance of our approach using transformer-based models \cite{DBLP:conf/nips/VaswaniSPUJGKP17}, regardless of being pretrained or fine-tuned using IsoMax+ loss.

\bibliographystyle{IEEEtran}
\bibliography{references}

\end{document}